\documentclass[10pt,twocolumn,letterpaper]{article}

\usepackage{wacv}      

\usepackage{graphicx}
\usepackage{amsmath}
\usepackage{amssymb}
\usepackage{booktabs}
\usepackage{algorithm}
\usepackage{algpseudocode}
\usepackage{multirow}
\usepackage[table, svgnames, dvipsnames]{xcolor}

\usepackage[pagebackref,breaklinks,colorlinks]{hyperref}

\usepackage[capitalize]{cleveref}
\crefname{section}{Sec.}{Secs.}
\Crefname{section}{Section}{Sections}
\Crefname{table}{Table}{Tables}
\crefname{table}{Tab.}{Tabs.}

\def\wacvPaperID{448} 
\def\confName{WACV}
\def\confYear{2026}
\def\method{AuViRe}

\begin{document}

\title{\method: Audio-visual Speech Representation Reconstruction for Deepfake Temporal Localization}

\author{Christos Koutlis\\
Information Technologies Institute @ CERTH\\
Thessaloniki, Greece\\
{\tt\small ckoutlis@iti.gr}
\and
Symeon Papadopoulos\\
Information Technologies Institute @ CERTH\\
Thessaloniki, Greece\\
{\tt\small papadop@iti.gr}
}
\maketitle

\begin{abstract}
With the rapid advancement of sophisticated synthetic audio-visual content, e.g., for subtle malicious manipulations, ensuring the integrity of digital media has become paramount. This work presents a novel approach to temporal localization of deepfakes by leveraging Audio-Visual Speech Representation Reconstruction (\method). Specifically, our approach reconstructs speech representations from one modality (e.g., lip movements) based on the other (e.g., audio waveform). Cross-modal reconstruction is significantly more challenging in manipulated video segments, leading to amplified discrepancies, thereby providing robust discriminative cues for precise temporal forgery localization. \method\ outperforms the state of the art by +8.9 AP@0.95 on LAV-DF, +9.6 AP@0.5 on AV-Deepfake1M, and +5.1 AUC on an in-the-wild experiment. Code available at \url{https://github.com/mever-team/auvire}.
\end{abstract}

\section{Introduction}
The advancement of generative AI has led to the rise of highly realistic synthetic media, often referred to as deepfakes \cite{mubarak2023survey}. These audio-visual manipulations threaten digital integrity, as they are designed to be perceptually convincing and capable of deceiving both humans and automated systems. This deceptive capacity fuels misinformation, erodes public trust, and carries significant societal and security risks. Ensuring the integrity of audio-visual content has become a critical challenge, driving the need for robust deepfake detection and temporal localization solutions.

While significant progress has been made in deepfake detection, accurately identifying and temporally localizing manipulations in audio-visual content remains a substantial challenge. Existing methods \cite{han2025towards,choi2024exploiting,yan2025generalizing,kashiani2025freqdebias,lin2024fake,xu2023tall,yao2023towards,shiohara2022detecting,chen2022self,larue2023seeable}, primarily focus on coarse-grained video classification instead of temporal forgery localization, while neglecting to also leverage the subtle cross-modal inconsistencies that are crucial for fine-grained temporal localization. In contrast, spatial forgery localization, being out of this work's scope, has been widely explored \cite{shuai2023locate,zhao2021multi}. Crucially, most state of the art multimodal deepfake detectors \cite{chugh2020not,cai2022you}, directly process the raw audio-visual signal instead of utilizing high-level features extracted by large models, which renders them prone to overfitting and less robust to content distortions, or utilize general purpose feature extractors \cite{cai2023glitch,zhang2023ummaformer} limiting their representational capacity.

\method\  addresses these limitations by explicitly modeling cross-modal discrepancies in speech representations. It leverages pre-trained speech-related feature extractors \cite{shi2022learning} for both visual and audio modalities, providing highly relevant and robust cues for detecting forged speech. Designed for Temporal Forgery Localization (TFL), our approach enables both fine-grained (frame-level) and coarse-grained (video-level) predictions rendering it also suitable for Deepfake Detection (DFD). Specifically, the proposed architecture first independently extracts visual and audio speech-related features. A representation reconstruction module then predicts visual speech representations from audio, complemented by unimodal (audio-audio and visual-visual) reconstructions. Finally, a reconstruction-discrepancy encoder processes these reconstruction errors to identify forgery features. \method\ outperforms the state of the art by +8.9 AP@0.95 on LAV-DF \cite{cai2022you}, +9.6 AP@0.5 on AV-Deepfake1M \cite{cai2024av} benchmarks. In addition, we provide a real-world implementation accounting for the presence of both visual and audio streams, large video lengths, and talking human presence, yielding +5.1 AUC on in-the-wild experiments. Our main contributions are:
\begin{enumerate}
    \item The proposed \method\ architecture, a novel framework for audio-visual deepfake temporal localization based on cross-modal reconstruction discrepancy modeling.
    \item An extensive evaluation on established benchmarks (LAV-DF, AV-Deepfake1M) achieving state of the art performance on both TFL and DFD.
    \item  A real-world evaluation of \method, demonstrating strong performance in the wild.
\end{enumerate}

\section{Related Work}
\textbf{Video Deepfake Detection.} Pertinent 
methods focus on spatio-temporal inconsistencies \cite{choi2024exploiting,han2025towards}, analyze frequency domain artifacts \cite{frank2020leveraging,kashiani2025freqdebias}, or use a multi-attentional approach \cite{zhao2021multi}. Approaches like SBI \cite{shiohara2022detecting} and SLADD \cite{chen2022self} use sophisticated augmentations to train on diverse synthetic forgeries, while SeeABLE \cite{larue2023seeable} reframes the problem as one-class anomaly detection. TALL transforms video clips into thumbnails, efficiently capturing spatio-temporal inconsistencies. Datasets like FaceForensics++ \cite{rossler2019faceforensics++}, Celeb-DF \cite{li2020celeb}, and DFDC \cite{dolhansky2020deepfake} have been instrumental in advancing these techniques. While achieving notable success in controlled environments, these video-level detectors struggle under partial multimodal forgeries.

\textbf{Multimodal Deepfake Detection.} The emergence of audio-visual deepfakes, where both visual and auditory streams are potentially manipulated, has necessitated the development of multimodal detection approaches. These methods look for inconsistencies across modalities, as maintaining perfect synchronization and consistency between visual and audio streams is still challenging for deepfake generators. Existing multimodal detectors such as those by Chugh et al. \cite{chugh2020not} and Cai et al. \cite{cai2022you}, typically concatenate or employ attention mechanisms to combine features extracted from raw audio and visual signals. However, directly processing raw signals can lead to overfitting to specific artifacts, reduced robustness to real-world distortions, and may not capture high-level semantic inconsistencies effectively. Some approaches use pre-extracted features, but rely on general-purpose encoders \cite{cai2023glitch,zhang2023ummaformer} lacking the capacity for speech-related understanding, which is crucial for the task. 

\textbf{Temporal Deepfake Localization.} Beyond binary classification (real vs. fake), the temporal localization of manipulated segments is important for forensic analysis, as it enables 
more fine-grained and transparent decisions about suspicious content. Recent works address this challenge, proposing methods that analyze spatio-temporal cross-modal consistency \cite{cai2023glitch,zhang2023ummaformer}. While these methods constitute an important step towards fine-grained deepfake analysis, they utilize general purpose feature extractors limiting their representational capacity. Our work addresses this by modeling frame-level cross-modal discrepancies in the speech latent space. DiMoDif \cite{koutlis2024dimodif} also considers speech-related features for cross-modal incongruity identification; however, our framework is fundamentally different leveraging cross-modal \textit{reconstruction discrepancy} instead of plain features encoding and comparison.

\section{Methodology}\label{sec:method}
\begin{figure}
    \centering
    \includegraphics[width=\linewidth,trim={5cm 8cm 4.5cm 7cm},clip]{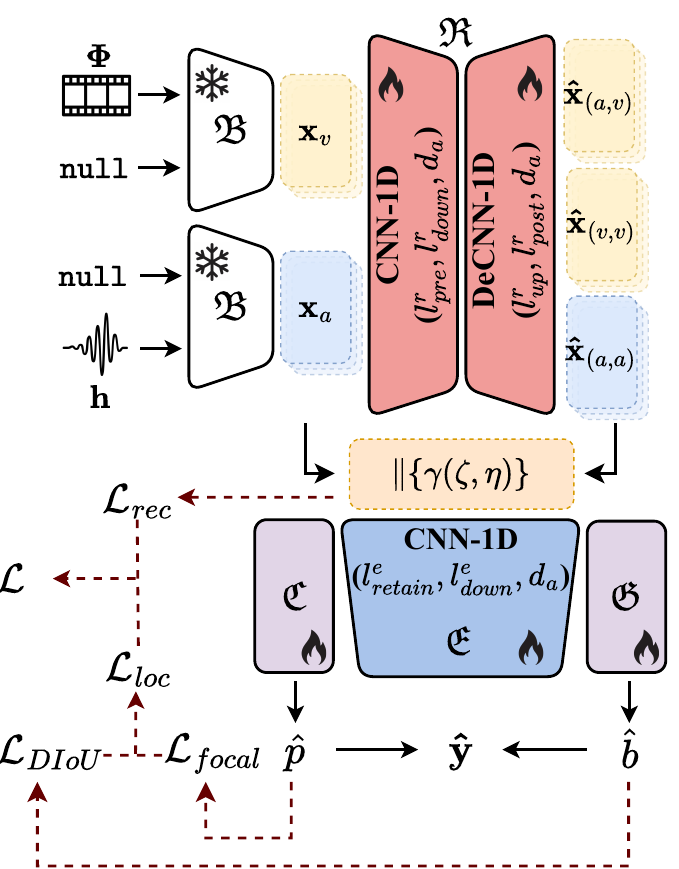}
    \caption{The proposed \method\ architecture.}
    \label{fig:auvire}
\end{figure}

In this section we elaborate on the proposed framework \method, which is illustrated in \cref{fig:auvire}.

\subsection{Problem Formulation}
Consider a dataset $\mathcal{V}$ comprising pairs of video and ground truth $(\mathbf{v},\mathbf{y})$. Each input video $\mathbf{v}$ has a duration of $\Delta$ seconds and consists of a visual stream $\boldsymbol{\Phi} \in \mathbb{R}^{t \times h \times w \times c}$ and an audio stream $\mathbf{h} \in \mathbb{R}^{s \times n}$. Specifically, $\boldsymbol{\Phi}$ contains $t$ frames, each of size $h \times w \times c$, while $\mathbf{h}$ contains $s$ samples across $n$ channels. The corresponding ground-truth annotation $\mathbf{y} \in \mathbb{R}^{t \times 3}$ is a concatenation of a per-frame manipulation target $p \in \{0,1\}^t$ and manipulation boundary information $b \in [0,\Delta]^{t \times 2}$. For each frame $\tau\in[0,t]$, $p^\tau=1$ indicates manipulation, and $b^\tau$ specifies the start and end times of the manipulated segment it belongs to, being valid for manipulated frames only. Our goal is to train a model $f(\mathbf{v})\rightarrow\mathbf{\hat{y}}$ to temporally localize the manipulated parts of deepfake videos.

\subsection{Visual and Audio Speech Representations}
Our architecture first extracts visual $\mathbf{x}_v\in\mathbb{R}^{t\times d}$ and audio $\mathbf{x}_a\in\mathbb{R}^{t\times d}$ representations from a self-supervised model \cite{shi2022learning}\footnote{We consider the AV-Hubert Base/LRS3/No-finetuning version from \url{https://facebookresearch.github.io/av_hubert/}.}, $\mathfrak{B}$, primarily designed for speech recognition, to incorporate multimodal speech information in the processing pipeline. Specifically, the visual stream $\boldsymbol{\Phi}$ and audio stream $\mathbf{h}$  of the video are processed separately by the backbone model avoiding modality blending that could eliminate fake samples' cross-modal discrepancies:
\begin{equation}
    \mathbf{x}_v = \mathfrak{B}(\boldsymbol{\Phi},\mathtt{null}),\quad\mathbf{x}_a = \mathfrak{B}(\mathtt{null},\mathbf{h})
\end{equation}
$\mathfrak{B}$ is a self-supervised model that uses separate audio and visual encoders feeding into a transformer to learn audio-visual speech representations by iteratively clustering multimodal features and predicting masked segments. In our framework $\mathfrak{B}$ is frozen during training.

\subsection{Representation Reconstruction Module}
The extracted representations $\mathbf{x}_v,\mathbf{x}_a$ are further processed by a reconstruction model, $\mathfrak{R}$, that utilizes them as input and target interchangeably. Its task is to predict the visual representation sequence based on the audio one as well as its visual-visual and audio-audio counterparts, namely:
\begin{equation}
    \mathbf{\hat{x}}_{(a,v)} = \mathfrak{R}(\mathbf{x}_a),\enspace
    \mathbf{\hat{x}}_{(v,v)} = \mathfrak{R}(\mathbf{x}_v),\enspace
    \mathbf{\hat{x}}_{(a,a)} = \mathfrak{R}(\mathbf{x}_a)
\end{equation}
with $\mathbf{\hat{x}}_{(\zeta,\eta)}\in\mathbb{R}^{t\times d}$; $\zeta,\eta\in\{a,v\}$, denoting the reconstruction of $\mathbf{x_\eta}$ from $\mathbf{x_\zeta}$. In general this process defines an easier objective when dealing with real samples that lack visual or audio artifacts. Thus, reconstruction discrepancies after subsequent processing should reveal manipulated video parts, if any. $\mathfrak{R}$ comprises a pre-projection block of $l^r_{pre}$, $d_{a}$-dimensional, 1D convolutional layers with kernel size $\kappa=3$ and stride $\sigma=1$, a down-sampling encoder-block of $l^r_{down}$, $d_{a}$-dimensional, 1D convolutional layers ($\kappa=\texttt{k},\sigma=2$), an up-sampling decoder-block of $l^r_{up}$, $d_{a}$-dimensional, 1D deconvolutional layers ($\kappa=\texttt{k},\sigma=2$), and a post-projection block of $l^r_{post}-1$, $d_{a}$-dimensional, and $1$, $d$-dimensional, 1D convolutional layers ($\kappa=3,\sigma=1$). Convolutional and deconvolutional layers are followed by Layer Normalization (LN) \cite{ba2016layer} and ReLU.

\subsection{Reconstruction-Discrepancy Encoder}
The reconstruction-discrepancy encoder module is employed to identify forgery features by amplifying the reconstruction discrepancies during manipulated video parts. Its input $\mathbf{x}\in\mathbb{R}^{t\times d\cdot 3}$ preparation includes reconstruction-discrepancy computation, and concatenation ($\|$):
\begin{align}\label{eq:encoder_input}
    \mathbf{x}=\gamma(a,v)\|\gamma(v,v)\|\gamma(a,a)
\end{align}
where $\gamma(\zeta,\eta)=\mathbf{\hat{x}}_{(\zeta,\eta)}-\mathbf{x}_\eta$. Then, $q$-dimensional forgery-specific features $\mathbf{f}\in\mathbb{R}^{t\times l\times q}$ are produced by the reconstruction-discrepancy encoder model, $\mathfrak{E}$:
\begin{align}
    \mathbf{f}=\mathfrak{E}(\mathbf{x})
\end{align}
implemented through a sequence of $l^e_{retain}$, $d_{a}$-dimensional, 1D convolutional layers ($\kappa=\texttt{k},\sigma=1$) retaining timesteps $t$ and a sequence of $l^e_{down}$, $d_{a}$-dimensional, 1D convolutional layers ($\kappa=\texttt{k},\sigma=2$) of downsampled timesteps (i.e., $t/2^{\lambda_{down}},\;\lambda_{down}=1,\dots,l^e_{down}$), additionally incorporating a feature pyramid network on top of its $l$ layers, with $l=l^e_{retain}+l^e_{down}$. Convolutional layers are followed by LN and ReLU.

\subsection{Predictions}
To obtain the temporal forgery localization predictions $\mathbf{\hat{y}}\in\mathbb{R}^{t\times l\times 3}$, our architecture uses a classification head $\mathfrak{C}$, for the manipulation detections $\hat{p}$ per frame $t$, and a regression head $\mathfrak{G}$, for the manipulation boundaries $\hat{b}$ if frame $t$ lies within a manipulated video part, by processing the forgery features $\mathbf{f}$:
\begin{equation}
    \hat{p} = \mathfrak{C}(\mathbf{f}),\quad \hat{b} = \mathfrak{G}(\mathbf{f})
\end{equation}
with $\mathbf{\hat{y}}=\hat{p}\|\hat{b}$, $\hat{p}\in\mathbb{R}^{t\times l}$, and $\hat{b}\in\mathbb{R}^{t\times l\times 2}$. $\mathfrak{C}$ and $\mathfrak{G}$ heads comprise 2, $d_{a}$-dimensional, and 1, $d_{out}$-dimensional, 1D convolutional layers ($\kappa=3,\sigma=1$) with $d_{out}=1$ for $\mathfrak{C}$ and $d_{out}=2$ for $\mathfrak{G}$, each followed by LN and ReLU. Additionally, SoftNMS \cite{bodla2017soft} is considered to enhance detection performance by  re-weighting overlapping boundary $\hat{b}$ predictions. Let $\mathcal{J} = \{ (s_j, e_j, \rho_j) \}_{j=1}^M$ be the set of $M$ predicted segments after applying SoftNMS, where $s_j$ and $e_j$ are the start and end times of the $j$-th segment, and $\rho_j$ is its confidence score. Finally, since \method\ is primarily trained on Temporal Forgery Localization (TFL) and not explicitly on binary deepfake detection, video-level predictions ($\hat{y}$) and targets ($y$) are post-processed as follows:
\begin{equation}\label{eq:dfd_target}
\hat{y} = \max_j \rho_j,\quad y = \bigvee_{\tau=1}^t p^\tau
\end{equation}
This allows us to evaluate \method\ on deepfake detection as an emergent property from its localization training.

\subsection{Objective Function}
To optimize our model, we consider a composite objective function accounting for manipulation detection and localization, as well as reconstruction-discrepancies minimization for real samples. Specifically, the focal loss $\mathcal{L}_{focal}$ \cite{lin2017focal} is considered for per-frame manipulation detection, the 
DIoU loss $\mathcal{L}_{DIoU}$ \cite{zheng2020distance} 
is considered for manipulation localization, and a mean absolute error-based loss $\mathcal{L}_{rec}$ is considered for reconstruction discrepancies minimization. The final loss $\mathcal{L}$ is computed with \cref{eq:loss}, after computing \cref{eq:loss_loc,eq:loss_rec}.
\begin{multline}\label{eq:loss_loc}
    \mathcal{L}_{loc} = \frac{1}{l}\sum_\lambda^l\sum_\tau^t\Big(\mathcal{L}_{focal}(\hat{p}^{\tau,\lambda},p^\tau)+\\
    p^\tau\cdot\mathcal{L}_{DIoU}(\hat{b}^{\tau,\lambda},b^\tau)
    \Big) \Big/ \sum_\tau^t p^\tau
\end{multline}
\begin{equation}\label{eq:loss_rec}
    \mathcal{L}_{rec} = \frac{\bigwedge_{\tau=1}^t \neg p^\tau}{t\cdot d}\sum_\tau^t\sum_\delta^d\sum_{\zeta,\eta\in\{a,v\}}\left|\mathbf{\hat{x}}_{\zeta,\eta}-\mathbf{x}_\eta\right|
\end{equation}
with $\zeta,\eta$ the pairs that define \cref{eq:encoder_input}.
\begin{equation}\label{eq:loss}
    \mathcal{L}=\frac{\mathcal{L}_{loc}+\mathcal{L}_{rec}}{2}
\end{equation}

\section{Experimental Setup}
\subsection{Datasets}
With respect to controlled experiments we consider two popular Temporal Forgery Localization benchmarks, i.e., LAV-DF \cite{cai2023glitch} and AV-Deepfake1M \cite{cai2024av}, which we also consider for video-level deepfake detection evaluation. LAV-DF contains 78,703 training, 31,501 validation, and 26,100 test samples. AV-Deepfake1M contains 746,180 training, 57,340 validation, and 343,240 test samples, the latter without released labels enabling evaluation only through Codabench\footnote{\url{https://deepfakes1m.github.io/2024/evaluation}}. Additionally, we conduct a real-world analysis 
on a collection of 371 videos curated with the help of professional fact-checkers, 
225 real and 146 manipulated, with median duration 1 minute (minimum: 10 sec, maximum: 11.3 minutes), spanning 39 languages\footnote{Identified via \url{https://speechbrain.github.io/}}.

\subsection{Implementation Details}
We trained our method for 100 epochs using the Adam optimizer with an initial learning rate of 0.001 (reduced on plateau) and a batch size of 64. We implemented early stopping with a patience of 10, and checkpointing based on the sum of validation AP@\{$\cdot$\} and AR@\{$\cdot$\}. We used a maximum sequence length ($t$) of 512, zero-padding smaller sequences. Key architectural parameters included a kernel size (\texttt{k}) of 15 and a model dimension ($d_a$) of 128. For the LAV-DF trained instance, the reconstruction model $\mathfrak{R}$ layers were configured as $l_{pre}^r=2$, $l_{down}^r=3$, $l_{up}^r=3$, and $l_{post}^r=2$, while the reconstruction-discrepancy encoder model $\mathfrak{E}$ layers were $l_{retain}^e=2$ and $l_{down}^e=2$. For the AV-Deepfake1M trained instance only  $l_{down}^r,l_{up}^r,l_{retain}^e,l_{down}^e=1$ deviate. We performed light tuning wrt $d_a\in\{32,64,128,256\}$, $l_{down}^r,l_{up}^r\in\{1,2,3\}$, and $l_{retain}^e,l_{down}^e\in\{1,2,3\}$. Training was conducted on NVIDIA GeForce RTX 4090 GPUs.

\subsection{Real-world Application Implementation}
The method implementation, as described in \cref{sec:method}, is well-suited for evaluation under controlled laboratory settings on academic benchmarks. However, deploying it to real-world web videos presents a multifaceted challenge. To address this, we apply several practical adaptations: we verify the existence of both visual and audio streams, implement video chunking\footnote{We consider 20 sec chunks.} for long content to mitigate memory issues, ensure the presence of adequately sized\footnote{Minimum face size is 3\% of frame area.} and speaking individuals\footnote{After empirical validation, we consider a mouth as non-talking if $\|\mathbf{x}_v^{\tau-1}-\mathbf{x}_v^{\tau}\|_2<2$, for a frame $\tau$.}, and apply the trained model only to relevant video segments of sufficient duration\footnote{Minimum segment duration is 2 sec.} (henceforth referred to as \texttt{valid} and denoted $\beta=\cup_i\beta_i\subseteq[0,\Delta]$) to obtain meaningful predictions. Beyond these adaptations, several unaddressed factors frequently trigger false positive predictions for small and sparse video segments, even in real videos. These include variations in video quality (e.g., compression rate, audio noise) and the inherent characteristics of the training dataset, which contains only short-duration manipulations. To counter these issues and robustly score videos for manipulation in the wild, we propose two methods for aggregating model predictions:

\begin{itemize}
    \item \textbf{Manipulation-fraction:} We define the manipulation-fraction score $\Psi_m$ as the ratio of the total length of predicted manipulated segments to the total length of \texttt{valid} video segments $\beta$.
    \begin{equation}
        \Psi_{m} = \frac{\mu\left(\bigcup_j \{ (s_j,e_j) \mid \rho_j > \theta \}\right)}{\mu\left(\bigcup_i \beta_i\right)}
        \label{eq:psi_m}
    \end{equation}
    For a probability threshold $\theta$, with $\mu$ denoting the Lebesgue measure.

    \item \textbf{Sweep-like:} This method, inspired by the plane sweep algorithm  of computational geometry domain \cite{de2000computational}, sorts all unique start $s_j$ and end $e_j$ points creating a sequence of event times $\iota_1 < \iota_2 < \ldots < \iota_Q$. For each sub-segment $[\iota_k, \iota_{k+1})$, we calculate the average confidence score across overlapping predicted intervals. The Sweep-like score $\Psi_s$ is then computed as the sum of these average scores, weighted by their respective segment durations:
    \begin{equation}
        \Psi_s = \sum_{k=1}^{Q-1} \left( \frac{\sum_{j \mid s_j \le \iota_k < e_j} \rho_j}{\Omega_k} \right) \cdot (\iota_{k+1} - \iota_k)
    \end{equation}
    where $\Omega_k = |\{ j \mid s_j \le \iota_k < e_j \}|$ denotes the number of active intervals in the segment $[\iota_k, \iota_{k+1})$. This approach effectively integrates the confidence scores over time, naturally accounting for overlaps by averaging scores in regions of intersection.
\end{itemize}

\begin{table*}[t]
    \centering
    \begin{tabular}{lllllllll}
    \toprule
        Method&Modality &  AP@0.5 & AP@0.75 & AP@0.95 & AR@100 &AR@50&AR@20&AR@10\\
    \midrule
        MDS \cite{chugh2020not} &$\mathcal{AV}$& 12.8& 1.6& 0.0& 37.9& 36.7& 34.4& 32.2\\
        AVFusion \cite{bagchi2021hear} &$\mathcal{AV}$& 65.4& 23.9& 0.1& 63.0& 59.3& 54.8& 52.1\\
        ActionFormer \cite{zhang2022actionformer} &$\mathcal{V}$& 95.3& 90.2& 23.7& 88.4& 89.6& 90.3& 90.4\\
        BA-TFD \cite{cai2022you}&$\mathcal{AV}$&76.9&38.5&0.3&66.9 &64.1 &60.8 &58.4\\
        BA-TFD+ \cite{cai2023glitch}&$\mathcal{AV}$&96.3&85.0&4.4&81.6 &80.5 &79.4 &78.8\\
        UMMAFormer \cite{zhang2023ummaformer}&$\mathcal{AV}$&\underline{98.8}&\underline{95.5}&\underline{37.6}&92.4&92.5&92.5&\underline{92.1} \\
        DiMoDif \cite{koutlis2024dimodif}&$\mathcal{AV}$&95.5&87.9&20.6&\underline{94.2}&\underline{93.7}&\underline{92.7}&91.4\\
        \midrule
        \method\ (ours)&$\mathcal{AV}$&\textbf{98.9}& \textbf{96.0}& \textbf{46.5}& \textbf{94.9}& \textbf{94.6}& \textbf{94.0}& \textbf{93.3}\\
    \bottomrule
    \end{tabular}
    \caption{Temporal forgery localization results on LAV-DF \cite{cai2022you}. Modality denotes the model's input type with $\mathcal{V}$ being visual and $\mathcal{A}$ audio. \textbf{Bold} indicates best and \underline{underline} second to best performance.}
    \label{tab:lavdf_tfl}
\end{table*}
\begin{table*}[t]
    \centering
    \resizebox{\textwidth}{!}{
    \begin{tabular}{lllllllllll}
    \toprule
        Method&Modality &  AP@0.5 & AP@0.75 & AP@0.9 & AP@0.95 & AR@50 &AR@30&AR@20&AR@10&AR@5\\
    \midrule
        MesoInception4 \cite{afchar2018mesonet}& $\mathcal{V}$& 08.50& 05.16& 01.89& 00.50& 39.27& 39.22& 39.00& 35.78& 24.59\\
        ActionFormer+VideoMAEv2 \cite{zhang2022actionformer,wang2023videomae}& $\mathcal{V}$& 20.24& 05.73& 00.57& 00.07& 19.97& 19.93& 19.81& 19.11& 17.80\\
        BA-TFD \cite{cai2022you}&$\mathcal{AV}$ &37.37 &6.34 &0.19 &0.02 &45.55 &40.37 &35.95 &30.66 &26.82 \\
        BA-TFD+ \cite{cai2023glitch}&$\mathcal{AV}$&44.42 &13.64 &0.48 &0.03 &48.86 &44.51 &40.37 &34.67 &29.88 \\
        UMMAFormer \cite{zhang2023ummaformer}&$\mathcal{AV}$ &51.64&28.07&7.65 &1.58 &44.07 &43.93 &43.45 &42.09 &40.27 \\
        DiMoDif \cite{koutlis2024dimodif}&$\mathcal{AV}$&\underline{86.93}&\underline{75.95}&\underline{28.72}&\underline{5.43}&\underline{81.57}&\underline{80.85}&\underline{80.25}&\underline{78.84}&\underline{76.64}\\
    \midrule
        \method\ (ours)&$\mathcal{AV}$&\textbf{96.5}&\textbf{89.3}&\textbf{42.9}&\textbf{11.7}&\textbf{86.0}&\textbf{85.8}&\textbf{85.5}&\textbf{84.9}&\textbf{83.8}\\
    \bottomrule
    \end{tabular}
    }
    \caption{Temporal forgery localization results on AV-Deepfake1M \cite{cai2024av}. Modality denotes the model's input type with $\mathcal{V}$ being visual and $\mathcal{A}$ audio. \textbf{Bold} indicates best and \underline{underline} second to best performance. *Reports validation performance.}
    \label{tab:avdf1m_tfl}
\end{table*}

\subsection{Evaluation}
We assess TFL performance using the standard AP@\{$\cdot$\} and AR@\{$\cdot$\} metrics. For Deepfake detection, we evaluate performance with the Area Under ROC Curve (AUC) and Average Precision (AP) metrics.

\section{Results}
\subsection{Temporal Forgery Localization}

\method\ was trained for TFL using the LAV-DF and AV-Deepfake1M datasets. As detailed in \cref{tab:lavdf_tfl,tab:avdf1m_tfl}, our method outperforms state-of-the-art approaches across all metrics on both test sets. Specifically, on LAV-DF dataset, \method\ shows a +8.9 absolute increase in terms of the challenging AP@0.95 metric, while maintaining strong performance on other metrics. On AV-Deepfake1M, \method\ demonstrates significant improvements: +9.6 AP@0.5, +13.4 AP@0.75, +14.2 AP@0.9, +6.3 AP@0.95, +4.5 AR@50, +5.0 AR@30, +5.3 AR@20, +6.1 AR@10, and +7.2 AR@5. These results, particularly outperforming DiMoDif \cite{koutlis2024dimodif} (which also uses speech-related inputs), highlight the significant value of our representation reconstruction module in constructing and leveraging distinctive forgery features.

\subsection{Video-level Deepfake Detection}
\begin{table}[t]
    \centering
    \begin{tabular}{lll}
    \toprule
        Method & Modality&AUC\\
    \midrule
    F\textsuperscript{3}-Net \cite{qian2020thinking}&$\mathcal{V}$&52.0\\
    MDS \cite{chugh2020not}&$\mathcal{AV}$&82.8\\
    EfficientViT \cite{coccomini2022combining}&$\mathcal{V}$&96.5\\
    BA-TFD \cite{cai2022you}&$\mathcal{AV}$&99.0\\
    UMMAFormer \cite{zhang2023ummaformer}&$\mathcal{AV}$&99.8\\
    DiMoDif \cite{koutlis2024dimodif} &$\mathcal{AV}$& \underline{99.84}\\
    \midrule
    \method\ (ours)&$\mathcal{AV}$&\textbf{99.94}\\
    \bottomrule
    \end{tabular}
    \caption{Video-level deepfake detection results on LAV-DF \cite{cai2022you}. Modality denotes the model's input type with $\mathcal{V}$ being visual and $\mathcal{A}$ audio. \textbf{Bold} indicates best and \underline{underline} second to best performance.}
    \label{tab:lavdf_dfd}
\end{table}

\begin{table}[t]
    \centering
    \begin{tabular}{lll}
    \toprule
        Method & Modality&AUC\\
    \midrule
    Video-LLaMA (13B) E5 \cite{zhang2023video}& $\mathcal{AV}$& 50.7\\
    LipForensics \cite{haliassos2021lips}&$\mathcal{V}$&51.6\\
    Face X-Ray \cite{li2020face}& $\mathcal{V}$& 61.5\\
    MesoInception4 \cite{afchar2018mesonet}&$\mathcal{V}$&50.1\\
    SBI \cite{shiohara2022detecting}&$\mathcal{V}$&65.8\\
    MDS \cite{chugh2020not}& $\mathcal{AV}$& 56.6\\
    DiMoDif \cite{koutlis2024dimodif} &$\mathcal{AV}$& \underline{96.3}\\
    \midrule
    \method\ (ours)&$\mathcal{AV}$&\textbf{99.8}\\
    \bottomrule
    \end{tabular}
    \caption{Video-level deepfake detection results on AV-Deepfake1M \cite{cai2024av}. Modality denotes the model's input type with $\mathcal{V}$ being visual and $\mathcal{A}$ audio. \textbf{Bold} indicates best and \underline{underline} second to best performance.}
    \label{tab:avdeepfake1m_dfd}
\end{table}
Even though \method\ is not trained for video-level deepfake detection, its frame-level outputs can be interpreted to provide this. Specifically, if the model does not localize any forgery, we consider it a non-detection. However, if any frame is detected as a deepfake, we interpret this as a positive deepfake detection for the video, as illustrated by \cref{eq:dfd_target}. \Cref{tab:lavdf_dfd,tab:avdeepfake1m_dfd} illustrate the performance of \method\ along with competitive methods for video-level deepfake detection on LAV-DF and AV-Deepfake1M, respectively. On LAV-DF, \method\ outperforms the previous state of the art, DiMoDif, achieving almost perfect performance, i.e., 99.94 AUC. On AV-Deepfake1M, \method\ outperforms the state of the art by +3.5 AUC with very strong performance, i.e., 99.8 AUC.

\subsection{Real-world Analysis}\label{subsec:real-world}
\begin{table}[t]
    \centering
    \begin{tabular}{ccc}
        \toprule
        method&AUC&AP\\
        \midrule
        DiMoDif \cite{koutlis2024dimodif} & 61.6 & 55.7 \\
        RealForensics \cite{haliassos2022leveraging} & 70.4 & 59.9\\
        \midrule
        \method\ ($\Psi_m$)&\textbf{75.5} & \textbf{67.2}\\
        \method\ ($\Psi_s$)&\underline{71.0} & \underline{62.3}\\
        \bottomrule
    \end{tabular}
    \caption{Real-world performance.}
    \label{tab:real_world_results}
\end{table}

\begin{table}[t]
    \centering
    \begin{tabular}{cccccc}
    \toprule
         && \multicolumn{2}{c}{$\Psi_m$}&\multicolumn{2}{c}{$\Psi_s$} \\
         \cmidrule(lr){3-4}\cmidrule(lr){5-6}
         Language&Videos (\#)& AUC&AP& AUC&AP\\
    \midrule
    English&282&77.9 & 64.4&73.6 & 60.0\\
    Welsh&42& 86.2 & 74.4&71.4 & 56.2\\
    French&24&60.1 & 68.3&54.5 & 59.7\\
    Greek&17&90.9 & 73.5&81.8 & 74.6\\
    Spanish&16&86.7 & 94.2&88.3 & 94.8\\
    \bottomrule
    \end{tabular}
    \caption{\method's real-world performance per language.}
    \label{tab:real_world_results_lang}
\end{table}

\Cref{tab:real_world_results} showcases \method's performance on a collection of real-world videos sourced from social media. We observed a significant performance drop compared to our controlled experiments on academic benchmarks (75.5 AUC vs. 99.9 AUC). This highlights the inherent difficulty of detecting ``in-the-wild'' manipulations, which often encompass diverse compression artifacts, varied lighting conditions, and subtle, sophisticated forgery techniques not typically present in curated datasets. Despite these challenges, our method outperforms existing state-of-the-art approaches, specifically DiMoDif and RealForensics, on the same real-world video set. This demonstrates \method's superior robustness and generalization capabilities in more realistic and uncontrolled environments. The manipulation-fraction variant $\Psi_m$ performs better than the sweep-like variant $\Psi_s$, which is attributed to the fact that \method's output contains several noisy predictions with many $\rho_i<0.01$, in this uncontrolled setting, that influence the sweep-like score. On the other hand, we set a threshold $\theta=0.01$ for the manipulation-fraction method to overcome this. To assess the performance of the real-world implementation on academic benchmarks we sample 1000 videos from LAV-DF's test set. \method\ achieves strong performance with 84.5 AUC - 93.7 AP for $\Psi_m$, and 96.2 AUC - 98.6 AP for $\Psi_s$, indicating that the absence of noisy outputs favors the sweep-like approach. Finally, we assess performance per language in this set and present the top 5 in terms of count in \cref{tab:real_world_results_lang}.
 
 \subsubsection{Performance-drop factors}
 Given the definition of $\Psi_m$ (\cref{eq:psi_m}) as the fraction of manipulated video parts and the distribution of $\Psi_m$ scores across real and fake samples (cf. Tab. \ref{tab:percentiles}), 
\begin{table}[b]
    
    \centering
    \begin{tabular}{ccccccc}
        \toprule
        $q$&5&10&25&50&75&90\\
        \midrule
        \textbf{real}&9.15&12.64& 21.60& 33.29& 46.87& 62.04 \\
        \textbf{fake}&23.27&27.55&  38.38& 53.28& 72.22& 84.21 \\
        \bottomrule
    \end{tabular}
    \caption{Percentiles of $\Psi_m$ distribution in \textbf{real} vs. \textbf{fake} samples.}
    \label{tab:percentiles}
\end{table}
the performance drop is primarily attributed to false positives rather than false negatives. Specifically, only 8 out of 146 \textbf{fake} videos obtain a $\Psi_m<23.27$, while 141 out of 225 \textbf{real} videos obtain a $\Psi_m>23.27$.
\begin{figure}[b]
    \centering
    \includegraphics[width=\linewidth]{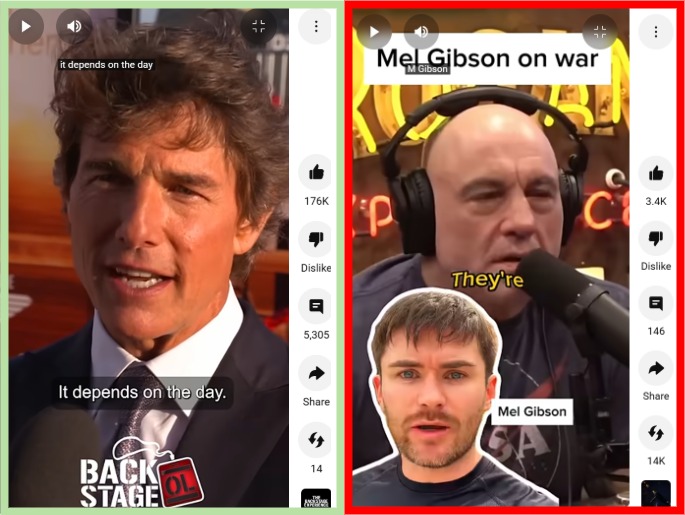}
    \caption{Examples of correctly (left; \url{https://www.youtube.com/shorts/xAPIjhkXF-0}) and erroneously (right; \url{https://www.youtube.com/shorts/D9mQGzG9dRk}) classified \textbf{real} samples.}
    \label{fig:samples}
\end{figure}
Fig. \ref{fig:samples} presents indicative examples of clearly correctly ($\Psi_m<12.64$) and erroneously ($\Psi_m>62.04$) classified \textbf{real} samples. The left (correct) example, \textit{in contrast to the right} (erroneous), is \textbf{free from mouth-occlusion}, contains \textbf{only one} clearly visible individual, has \textbf{no hards cuts}, and presents clear audio speech \textbf{without mix with music}; interestingly it exhibits some low-level background noise that does not affect the result. After visual inspection of all clearly correct ($\Psi_m<12.64$) and erroneous ($\Psi_m>62.04$) predictions for \textbf{real} samples we identified several underlying factors that potentially affect performance. Tab. \ref{tab:factors} illustrates the corresponding fractions, clearly showing that significantly higher values are present in misclassified videos. The highest differences, indicating the most prominent factors of performance drop, are found for videos containing \textit{long shots}, \textit{multiple subjects that talk during the course of the video}, \textit{multi-subject frames}, and \textit{non-English speech}. In contrast, \textit{monochromatism}, \textit{gender}, \textit{mouth-occlusion}, and \textit{race} factors were found to be less significant.
\begin{table}[b]
\caption{Factors' fractions for \textbf{real} videos obtaining correct (low $\Psi_m$) versus erroneous (high $\Psi_m$) predictions.}
\label{tab:factors}
\centering
\resizebox{\linewidth}{!}{
\begin{tabular}{lp{2cm}p{2.3045cm}p{2.1cm}}
\toprule
& \textbf{Non-English Speech} & \textbf{Non-Native Accent} & \textbf{Speech/Music Mix} \\
\midrule
\textbf{Corr. (Low $\Psi_m$)} & 13.04\% & 4.35\% & 17.39\% \\
\textbf{Err. (High $\Psi_m$)} & 34.78\% & 17.39\% & 26.09\% \\
\midrule\midrule
& \textbf{Low Fidelity} & \textbf{Mono-chromatic} & \textbf{Long Shot} \\
\midrule
\textbf{Corr. (Low $\Psi_m$)} & 0.00\% & 4.35\% & 4.35\% \\
\textbf{Err. (High $\Psi_m$)} & 13.04\% & 4.35\% & 43.48\% \\
\midrule\midrule
& \textbf{Multiple Subjects} & \textbf{Multi-Subject Frames} & \textbf{Hard Cuts} \\
\midrule
\textbf{Corr. (Low $\Psi_m$)} & 13.04\% & 0.00\% & 4.35\% \\
\textbf{Err. (High $\Psi_m$)} & 43.48\% & 30.43\% & 13.04\% \\
\midrule\midrule
& \textbf{Mouth Occl.} & \textbf{Non-Caucasian Speaker} & \textbf{Female Speaker} \\
\midrule
\textbf{Corr. (Low $\Psi_m$)} & 17.39\% & 13.04\% & 17.39\% \\
\textbf{Err. (High $\Psi_m$)} & 21.74\% & 21.74\% & 17.39\% \\
\bottomrule
\end{tabular}
}
\end{table}
To mitigate these challenges, future work should focus on the inclusion of language-diverse training samples and more accurate isolation of each individual's visuals and audio during inference. 

\subsection{Robustness}
\begin{figure*}
    \centering
    \includegraphics[width=\linewidth,trim={0cm 1.5cm 0cm 3.5cm},clip]{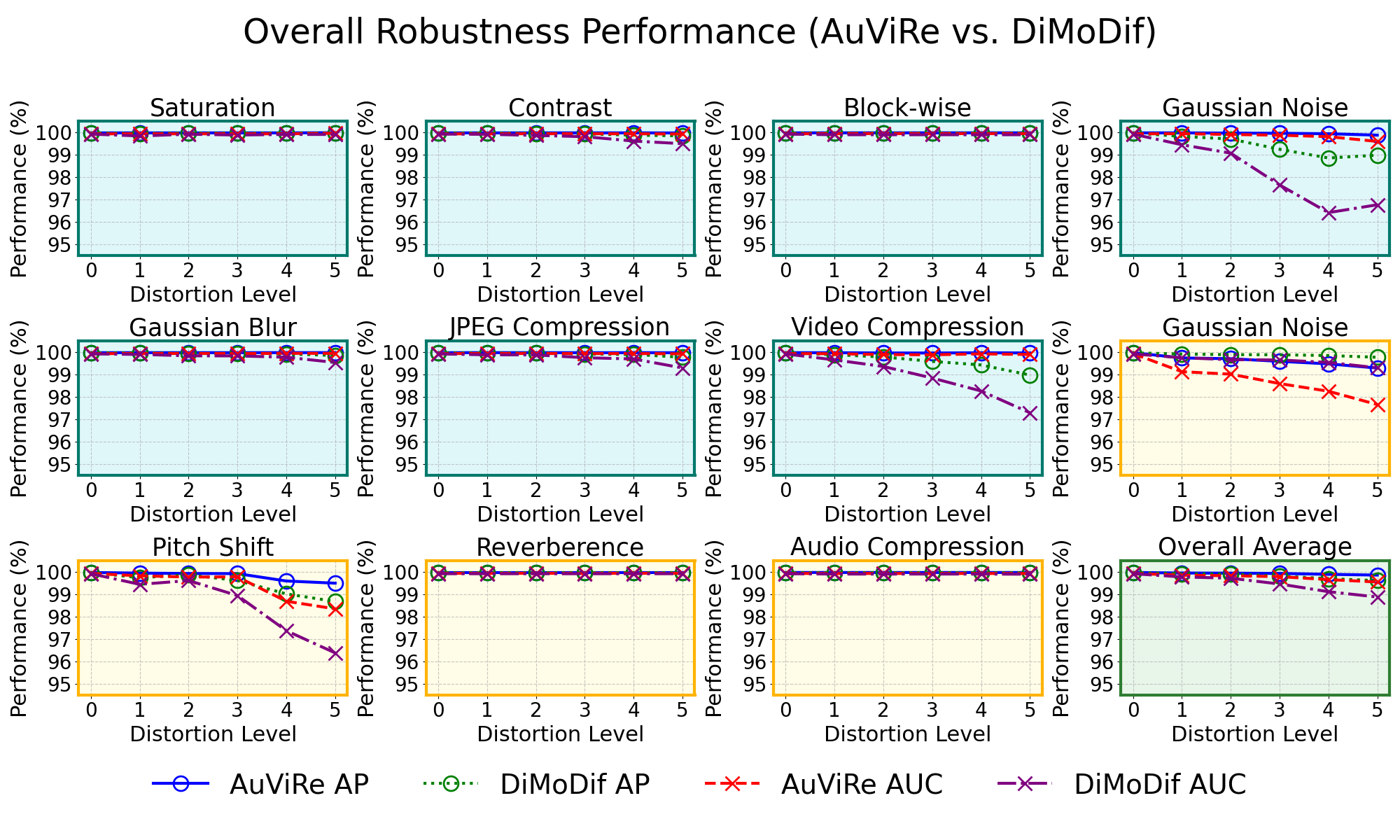}
    \caption{Robustness analysis using visual (cyan plots) and audio (yellow plots) distortions. Overall average robustness is also reported.}
    \label{fig:robustness}
\end{figure*}

\cref{fig:robustness} compares the robustness of \method\ against the next most competitive approach, DiMoDif. The evaluation was conducted on 1,000 unseen samples from the LAV-DF dataset, incorporating seven visual and four audio distortions, each applied across five intensity levels. Both models exhibit high robustness to most distortions, a quality partly attributed to the resilience of their underlying speech-related features. DiMoDif's performance degrades most significantly under visual distortions, specifically Gaussian noise and video compression, and with the pitch shift audio distortion. In contrast, \method\ remains resilient to these effects. Conversely, \method\ maintains near-perfect robustness across all conditions except for Gaussian audio noise, where it experiences a more substantial performance drop than DiMoDif. 


\subsection{Ablation study}
\begin{table*}
\centering
\caption{Ablation analysis on LAV-DF for loss composition ($\mathcal{L}$ comp.), reconstruction-discrepancy encoder input ($\mathfrak{E}$ input), reconstruction-discrepancy operation ($\gamma(\zeta,\eta)$), backbone ($\mathfrak{B}$), and model type ($\mathcal{M}$). AP@\{0.5,0.75,0.95\} and AR\{100,50,20,10\} metrics are reported along with corresponding rank in parenthesis. Last column reports per instance average rank ($\bar{\text{R}}$).}
\label{tab:ablation}
\resizebox{\textwidth}{!}{
\begin{tabular}{llllllllllr}
\toprule
 & Compon. & Config. & AP@0.5 & AP@0.75 & AP@0.95 & AR@100 & AR@50 & AR@20 & AR@10 & $\bar{\text{R}}$ \\
\midrule
0 & - & \method\ & \textbf{98.91} (1) & \textbf{96.03} (1) & 46.50 (3) & 94.93 (2) & 94.56 (3) & 93.99 (3) & 93.27 (3) & \textbf{2.2} \\
 \midrule
1 & \multirow{7}{*}{$\mathcal{L}$ comp.} & \{$\mathcal{L}_{focal}$,$\mathcal{L}_{DIoU}$\} & 98.63 (4) & 94.33 (12) & 43.55 (7) & 94.09 (16) & 93.59 (16) & 92.80 (17) & 91.78 (17) & 12.1 \\
2 &  & \{$\mathcal{L}_{focal}$,$\mathcal{L}_{s-L1}$\} & 98.03 (15) & 93.63 (15) & 36.36 (14) & 94.92 (3) & 94.56 (4) & 93.99 (4) & 93.12 (4) & 9.2 \\
3 &  & \{$\mathcal{L}_{focal}$,$\mathcal{L}_{s-L1}$,$\mathcal{L}_{rec}$\} & 97.97 (17) & 93.78 (14) & 35.52 (15) & 94.82 (6) & 94.58 (2) & \textbf{94.12} (1) & 93.28 (2) & 9.0 \\
4 &  & \{$\mathcal{L}_{focal}$,$\mathcal{L}_{DIoU}$,$\mathcal{L}_{det}$\} & 98.47 (9) & 94.60 (11) & \textbf{48.68} (1) & 94.13 (15) & 93.76 (15) & 93.05 (16) & 92.17 (16) & 11.2 \\
5 &  & \{$\mathcal{L}_{focal}$,$\mathcal{L}_{s-L1}$,$\mathcal{L}_{det}$\} & 98.08 (14) & 91.70 (19) & 26.91 (20) & 94.72 (9) & 94.41 (7) & 93.70 (9) & 92.68 (10) & 13.2 \\
6 &  & \{$\mathcal{L}_{focal}$,$\mathcal{L}_{DIoU}$,$\mathcal{L}_{det}$,$\mathcal{L}_{rec}$\} & 98.18 (11) & 94.69 (9) & 48.65 (2) & 94.49 (14) & 94.06 (13) & 93.40 (13) & 92.64 (12) & 10.2 \\
7 &  & \{$\mathcal{L}_{focal}$,$\mathcal{L}_{s-L1}$,$\mathcal{L}_{det}$,$\mathcal{L}_{rec}$\} & 98.28 (10) & 92.89 (16) & 32.26 (19) & \textbf{95.00} (1) & \textbf{94.64} (1) & 93.94 (5) & 92.97 (8) & 9.4 \\
 \midrule
8 & \multirow{10}{*}{$\mathfrak{E}$ input} & \{$(a,v)$\} & 98.00 (16) & 94.70 (8) & 38.12 (13) & 94.60 (11) & 94.23 (10) & 93.63 (11) & 92.65 (11) & 11.5 \\
9 &  & \{$(v,a)$\} & 97.65 (19) & 93.79 (13) & 41.90 (10) & 94.60 (10) & 94.10 (12) & 93.42 (12) & 92.61 (13) & 12.9 \\
10 &  & \{$(a,a)$\} & 95.88 (20) & 88.95 (20) & 34.78 (16) & 93.42 (18) & 92.88 (19) & 92.07 (19) & 91.12 (19) & 18.7 \\
11 &  & \{$(v,v)$\} & 83.38 (21) & 68.00 (21) & 13.25 (21) & 83.75 (21) & 82.42 (21) & 80.02 (21) & 77.47 (21) & 21.0 \\
12 &  & \{$(a,v)$,$(v,a)$\} & 98.53 (8) & 94.67 (10) & 42.52 (9) & 94.50 (13) & 93.91 (14) & 93.07 (14) & 92.17 (15) & 11.5 \\
13 &  & \{$(a,v)$,$(a,a)$\} & 98.59 (7) & 94.99 (5) & 42.93 (8) & 94.58 (12) & 94.19 (11) & 93.71 (8) & 93.08 (6) & 8.0 \\
14 &  & \{$(a,v)$,$(v,v)$\} & 98.17 (12) & 94.88 (6) & 43.98 (6) & 94.75 (8) & 94.39 (8) & 93.78 (7) & 92.99 (7) & 7.8 \\
15 &  & \{$(a,v)$,$(v,a)$,$(a,a)$\} & 98.78 (3) & 95.31 (3) & 44.35 (4) & 94.77 (7) & 94.28 (9) & 93.65 (10) & 92.90 (9) & 6.0 \\
16 &  & \{$(a,v)$,$(v,a)$,$(v,v)$\} & 98.81 (2) & 95.57 (2) & 41.68 (11) & 94.83 (4) & 94.43 (6) & 93.87 (6) & 93.09 (5) & 5.1 \\
17 &  & \{$(a,v)$,$(v,a)$,$(a,a)$,$(v,v)$\} & 98.62 (5) & 94.76 (7) & 44.31 (5) & 94.82 (5) & 94.49 (5) & 94.05 (2) & \textbf{93.49} (1) & 4.5 \\
 \midrule
18 & $\gamma(\zeta,\eta)$ & $\mathbf{\hat{x}}_{(\zeta,\eta)}\cdot\mathbf{x}_\eta$ & 98.61 (6) & 95.14 (4) & 39.75 (12) & 93.84 (17) & 93.54 (17) & 93.06 (15) & 92.39 (14) & 11.5 \\
 \midrule
19 & $\mathfrak{B}$ & (Ma et al. 2022) \cite{ma2022visual} & 97.71 (18) & 92.11 (17) & 33.89 (18) & 92.85 (20) & 92.41 (20) & 91.83 (20) & 91.10 (20) & 18.8 \\
 \midrule
20 & $\mathcal{M}$ & Transformer & 98.09 (13) & 91.93 (18) & 34.55 (17) & 93.38 (19) & 92.98 (18) & 92.27 (18) & 91.50 (18) & 17.1 \\
\bottomrule
\end{tabular}
}
\end{table*}
\cref{tab:ablation} reports ablation experiments on the LAV-DF dataset, investigating the loss composition ($\mathcal{L}$ comp.), reconstruction-discrepancy encoder input ($\mathfrak{E}$ input), reconstruction-discrepancy operation ($\gamma(\zeta,\eta)$), backbone ($\mathfrak{B}$), and model type ($\mathcal{M}$). \method's original configuration includes $\{\mathcal{L}_{focal},\mathcal{L}_{DIoU},\mathcal{L}_{rec}\}$ for $\mathcal{L}$ composition, $\{(a,v),(a,a),(v,v)\}$ for $\mathfrak{E}$ input, $(\mathbf{\hat{x}}_{(\zeta,\eta)}-\mathbf{x}_\eta)$ for $\gamma(\zeta,\eta)$, Shi et al. (2022) \cite{shi2022learning} for $\mathfrak{B}$, and CNN for $\mathcal{M}$. We also consider smooth L1 loss ($\mathcal{L}_{s-L1}$) \cite{girshick2015fast} as an alternative to $\mathcal{L}_{DIoU}$ for localization optimization, and video-level binary cross-entropy ($\mathcal{L}_{det}$) for auxiliary supervision. The proposed configuration achieves the highest AP@\{0.5,0.75\} scores and the best average rank. Immediate competitors involve slight modifications to $\mathfrak{E}$'s input. In contrast, significant modifications, e.g., using only $\{(v,v)\}$, lead to the worst performance highlighting the critical importance of cross-modal representation reconstruction. Limited variability is observed wrt loss composition though certain configurations show occasional deviations. Finally, using multiplication as discrepancy estimator, the backbone from Ma et al. \cite{ma2022visual}, or Transformer\footnote{Identical configuration, except the downsampling/upsampling blocks are Transformer encoder blocks with conv./deconv. final layers.} model type yield suboptimal performance.

\subsection{Computational Efficiency}
In addition to achieving state of the art performance across both controlled and in-the-wild environments, \method\ exhibits notable computational efficiency. Across the 371 videos comprising our real-world study (cf. \cref{subsec:real-world}), the average $\texttt{processing-time}/\texttt{video-duration}$ ratio was computed to be 0.58 with a standard deviation of 0.25, demonstrating faster than real-time processing capability. This contrasts with methods like DiMoDif, which achieve real-time operation (x1.0 \texttt{video-duration}). In addition, \method\ has 12.1M learnable parameters requiring 1.0 GFLOP for a forward pass of a single video features, and processes $\sim$43 FPS (including feature extraction and simultaneous audio processing).

\section{Conclusions}
We introduced \method, a novel method that leverages discrepancies in reconstructed latent space speech representations to identify deepfakes. \method\ works by first extracting speech representations from both visual and audio modalities. These are then reconstructed using information from either the same modality or its counterpart. The core of our approach lies in processing the reconstruction discrepancy with an encoder, which effectively amplifies errors in manipulated video segments. This allows \method\ to precisely pinpoint deepfake alterations. Our extensive experiments demonstrate that \method\ not only outperforms state-of-the-art methods but also exhibits strong robustness, generalization, and practical applicability in real-world scenarios. Notably, although not explicitly trained for video-level audio-visual deepfake detection, it surpasses previous state of the art based on its transformed localization output.

\section{Acknowledgments}
This work is partially funded by the Horizon Europe projects AI4TRUST (GA No. 101070190) and AI-CODE (GA No. 101135437).

{\small
\bibliographystyle{ieee_fullname}
\bibliography{egbib}
}
\clearpage
\appendix
\section{Hyperparameter tuning}
\begin{figure*}[!h]
    \centering
    \includegraphics[width=\textwidth]{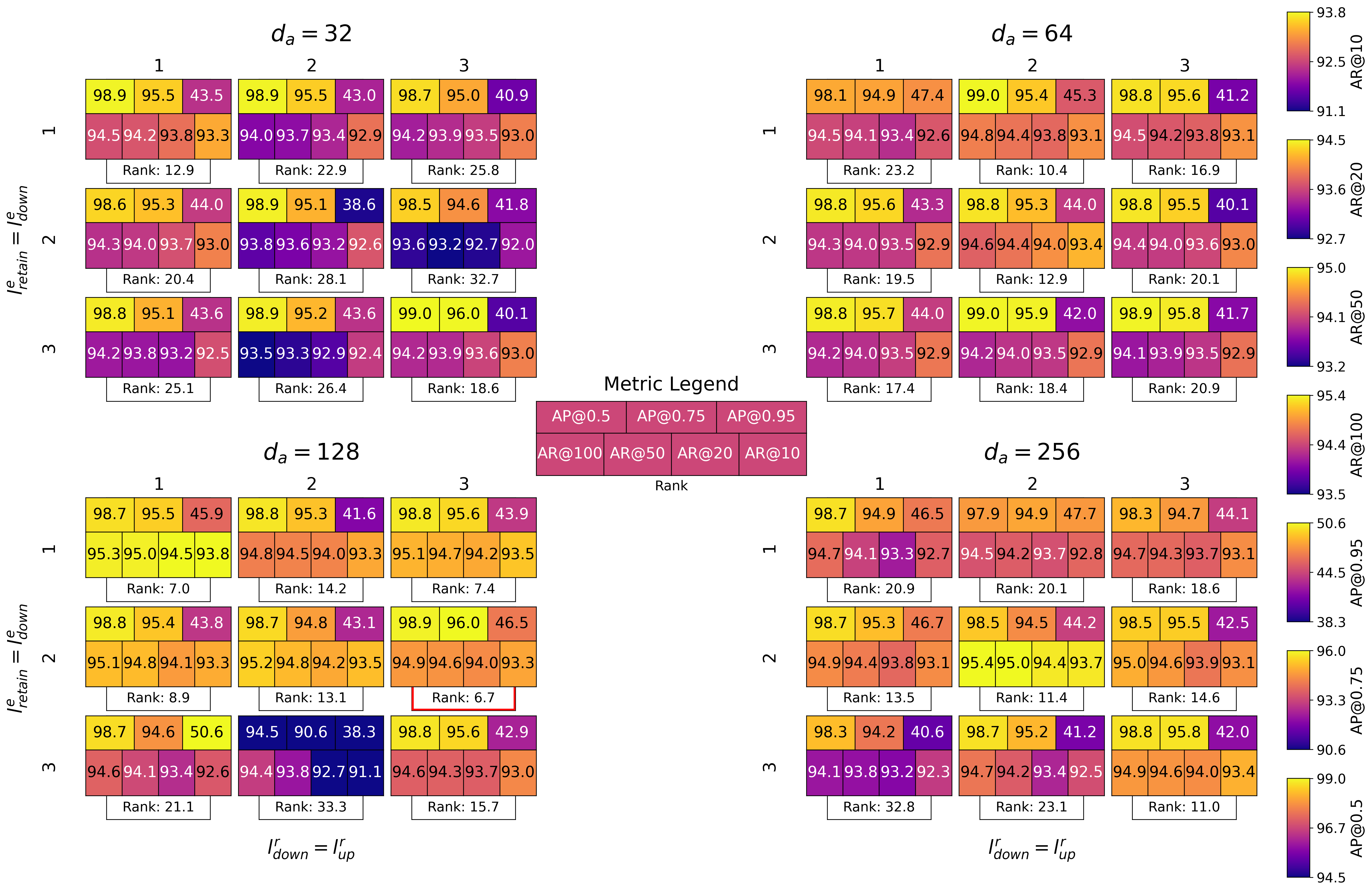}
    \caption{Hyperparameter grid results on LAV-DF. Best rank is highlighted with red.}
    \label{fig:tuning_lavdf}
\end{figure*}

\begin{figure*}[!h]
    \centering
    \includegraphics[width=\textwidth]{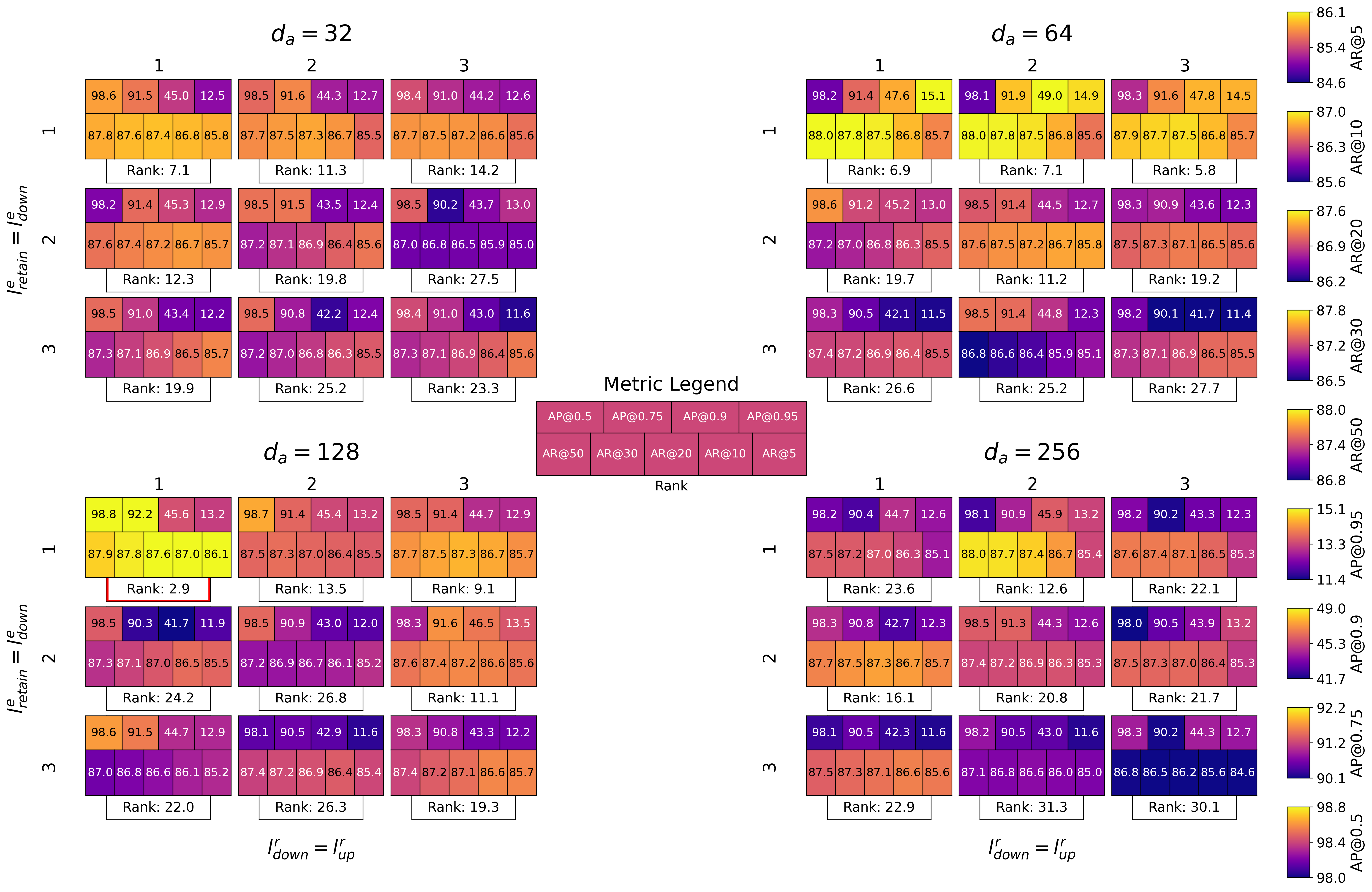}
    \caption{Hyperparameter grid results on AV-Deepfake1M. Best rank is highlighted with red.}
    \label{fig:tuning_avdf1m}
\end{figure*}

\cref{fig:tuning_lavdf} presents our hyperparameter tuning experimental results. These are obtained by training and evaluating \method\ on LAV-DF on a grid of 36 experiments defined by $d_a\in\{32,64,128,256\}$, $l_{down}^r=l_{up}^r\in\{1,2,3\}$, and $l_{retain}^e=l_{down}^e\in\{1,2,3\}$. High AP is achieved with $d_a\in\{32,64\}$ but with lower AR scores. $d_a\in\{128,256\}$ strike a balance between AP and AR with the best configuration obtaining average rank 6.7 defined by $d_a=128$, $l_{down}^r=l_{up}=3$, $l_{retain}^e=l_{down}^e=2$. The same analysis on AV-Deepfake1M (validation set) is illustrated in \cref{fig:tuning_avdf1m}.

\section{Robustness}
\cref{tab:backbone_robustness} reports average backbone $\mathfrak{B}$ feature similarity among original and distorted versions of the same LAV-DF samples, showcasing its high resilience that contributes to \method's robustness.

\begin{table*}[t]
\centering
\caption{Average cosine similarity of $\mathfrak{B}$ extracted features, between original and distorted versions of 1,000 LAV-DF unseen samples.}
\label{tab:backbone_robustness}
\begin{tabular}{llccccc}
\toprule
Modality & Distortion & Level 1 & Level 2 & Level 3 & Level 4 & Level 5 \\
\midrule
\multirow{7}{*}{Visual} & Color Saturation & 0.96 $\pm$ 0.05 & 0.96 $\pm$ 0.05 & 0.97 $\pm$ 0.05 & 0.96 $\pm$ 0.05 & 0.97 $\pm$ 0.05 \\
 & Color Contrast & 0.95 $\pm$ 0.05 & 0.93 $\pm$ 0.06 & 0.89 $\pm$ 0.08 & 0.83 $\pm$ 0.10 & 0.73 $\pm$ 0.10 \\
 & Block-wise & 0.96 $\pm$ 0.04 & 0.96 $\pm$ 0.04 & 0.94 $\pm$ 0.06 & 0.92 $\pm$ 0.08 & 0.88 $\pm$ 0.11 \\
 & Gaussian Noise & 0.87 $\pm$ 0.08 & 0.79 $\pm$ 0.10 & 0.64 $\pm$ 0.13 & 0.48 $\pm$ 0.16 & 0.09 $\pm$ 0.10 \\
 & Gaussian Blur & 0.94 $\pm$ 0.06 & 0.92 $\pm$ 0.08 & 0.86 $\pm$ 0.10 & 0.79 $\pm$ 0.11 & 0.71 $\pm$ 0.13 \\
 & JPEG & 0.96 $\pm$ 0.05 & 0.94 $\pm$ 0.07 & 0.90 $\pm$ 0.08 & 0.84 $\pm$ 0.09 & 0.75 $\pm$ 0.10 \\
 & Video Compression & 0.89 $\pm$ 0.06 & 0.85 $\pm$ 0.06 & 0.77 $\pm$ 0.09 & 0.67 $\pm$ 0.09 & 0.59 $\pm$ 0.10 \\
\midrule
\multirow{4}{*}{Audio} & Gaussian Noise & 0.81 $\pm$ 0.07 & 0.73 $\pm$ 0.09 & 0.66 $\pm$ 0.10 & 0.58 $\pm$ 0.11 & 0.49 $\pm$ 0.11 \\
 & Pitch Shift & 0.88 $\pm$ 0.04 & 0.67 $\pm$ 0.18 & 0.49 $\pm$ 0.25 & 0.35 $\pm$ 0.22 & 0.26 $\pm$ 0.20 \\
 & Reverberence & 1.00 $\pm$ 0.03 & 1.00 $\pm$ 0.03 & 1.00 $\pm$ 0.04 & 0.99 $\pm$ 0.04 & 0.99 $\pm$ 0.06 \\
 & Audio Compression & 0.88 $\pm$ 0.17 & 0.60 $\pm$ 0.10 & 0.51 $\pm$ 0.11 & 0.56 $\pm$ 0.19 & 0.99 $\pm$ 0.06 \\
\bottomrule
\end{tabular}
\end{table*}

\section{Generalization}
\cref{tab:generalization_dfd,tab:generalization_tfl} show cross-dataset generalization performance for DFD and TFL. Trained on AV-Deepfake1M, a set ten times larger with higher quality generated content, the model achieves strong performance on LAV-DF, with 93.33 AUC (+7.03 compared to DiMoDif) and 43.3 AP@0.75 (+17.1 compared to DiMoDif). Trained on LAV-DF \method\ performs on par to DiMoDif (-1.51 AUC, +2.1 AP@0.75). The failure of both models at 90/95\% IoU is attributed to the x2 larger forged segments of LAV-DF vs. AV-Deepfake1M. 

\begin{table*}[t]
\centering
\caption{Generalization performance on Deepfake Detection task.}
\label{tab:generalization_dfd}
\begin{tabular}{llcccc}
\toprule
&&\multicolumn{4}{c}{Tested on}\\
\cmidrule(lr){3-6}
& & \multicolumn{2}{c}{LAV-DF} & \multicolumn{2}{c}{AVD1M} \\
\cmidrule(lr){3-4} \cmidrule(lr){5-6}
Trained on& Model& AP & AUC & AP & AUC \\
\midrule
\multirow{2}{*}{AVD1M} & DiMoDif & 94.98 & 86.30 & \color{gray}- & \color{gray}96.30 \\
& AuViRe & \textbf{97.66} & \textbf{93.33} & \textbf{\color{gray}-} & \textbf{\color{gray}99.78} \\

\midrule
\multirow{2}{*}{LAV-DF} & DiMoDif & \color{gray}99.94 & \color{gray}99.84 & - & \textbf{67.21} \\
& AuViRe & \textbf{\color{gray}99.98} & \textbf{\color{gray}99.94} & - & 65.70 \\
\bottomrule
\end{tabular}%
\end{table*}

\begin{table*}[t] 
  \centering
  \caption{Generalization performance on Temporal Forgery Localization task.}
  \label{tab:generalization_tfl}
  \begin{tabular}{lll cccc cccccc} 
    \toprule
     &  & & \multicolumn{4}{c}{AP} & \multicolumn{5}{c}{AR} \\
    \cmidrule(lr){4-7} \cmidrule(lr){8-13}
    Trained on& Tested on& Method& @0.5 & @0.75 & @0.9 & @0.95 & @100 & @50 & @30 & @20 & @10 & @5 \\
    \midrule
    \multirow{2}{*}{AVD1M} & \multirow{2}{*}{LAV-DF} & DiMoDif & \textbf{59.8} & 26.2 & 2.3 & 0.2 & 75.3 & 72.3 & 69.7 & 67.3 & 62.5 & 56.2 \\
    & & AuViRe & 53.8 & \textbf{43.3} & \textbf{13.9} & \textbf{0.9} & \textbf{78.0} & \textbf{76.5} & \textbf{75.2} & \textbf{74.1} & \textbf{71.4} & \textbf{66.8} \\
    \midrule
    \multirow{2}{*}{LAV-DF} & \multirow{2}{*}{AVD1M} & DiMoDif & \textbf{15.9} & 4.3 & 0.3 & 0.02 & - & 28.9 & \textbf{27.3} & \textbf{25.7} & \textbf{22.5} & \textbf{19.1} \\
    & & AuViRe & 14.7 & \textbf{6.4} & \textbf{0.6} & \textbf{0.05} & -&\textbf{29.3} & 26.6 & 24.6 & 21.3 & 18.2 \\
    \bottomrule
  \end{tabular}
\end{table*}

\section{Threshold Calibration}
\begin{figure*}
    \centering
    \includegraphics[width=\linewidth]{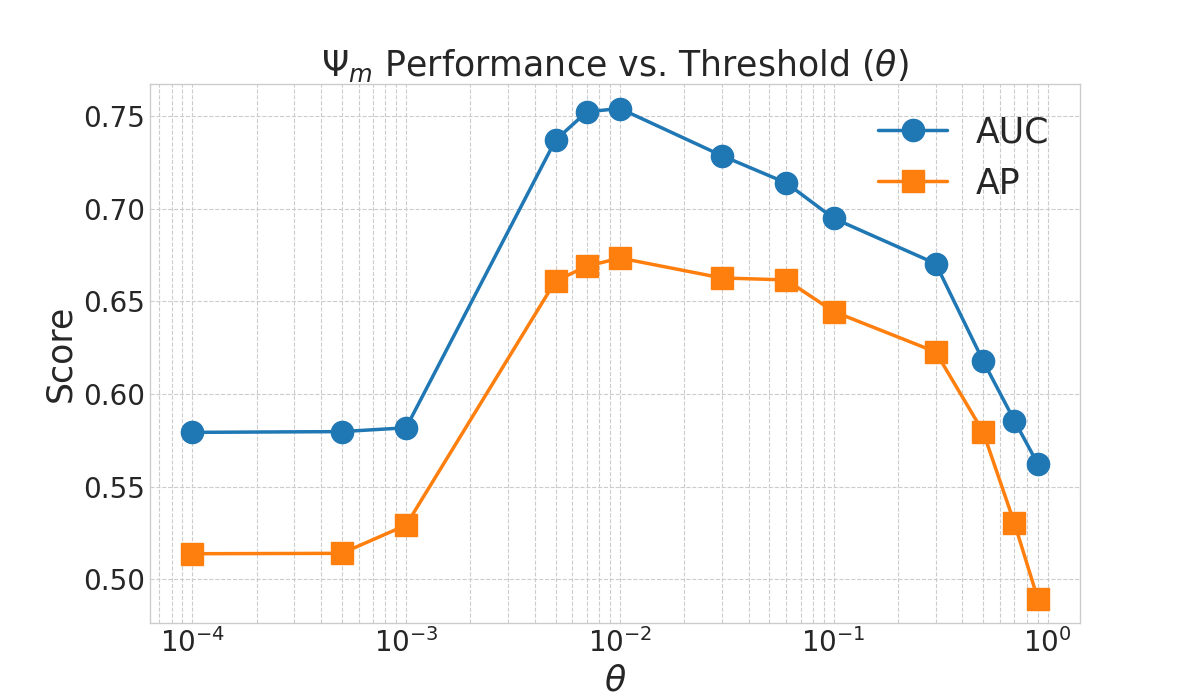}
    \caption{Calibration of $\Psi_m$'s $\theta$ parameter.}
    \label{fig:calibration}
\end{figure*}

Fig. \ref{fig:calibration}, presents a systematic calibration analysis using the real-world data indicating 0.01 as $\Psi_m$'s optimal probability threshold $\theta$.

\section{Real-world performance drop: Examples}
\begin{figure*}
    \centering
    \includegraphics[width=\textwidth]{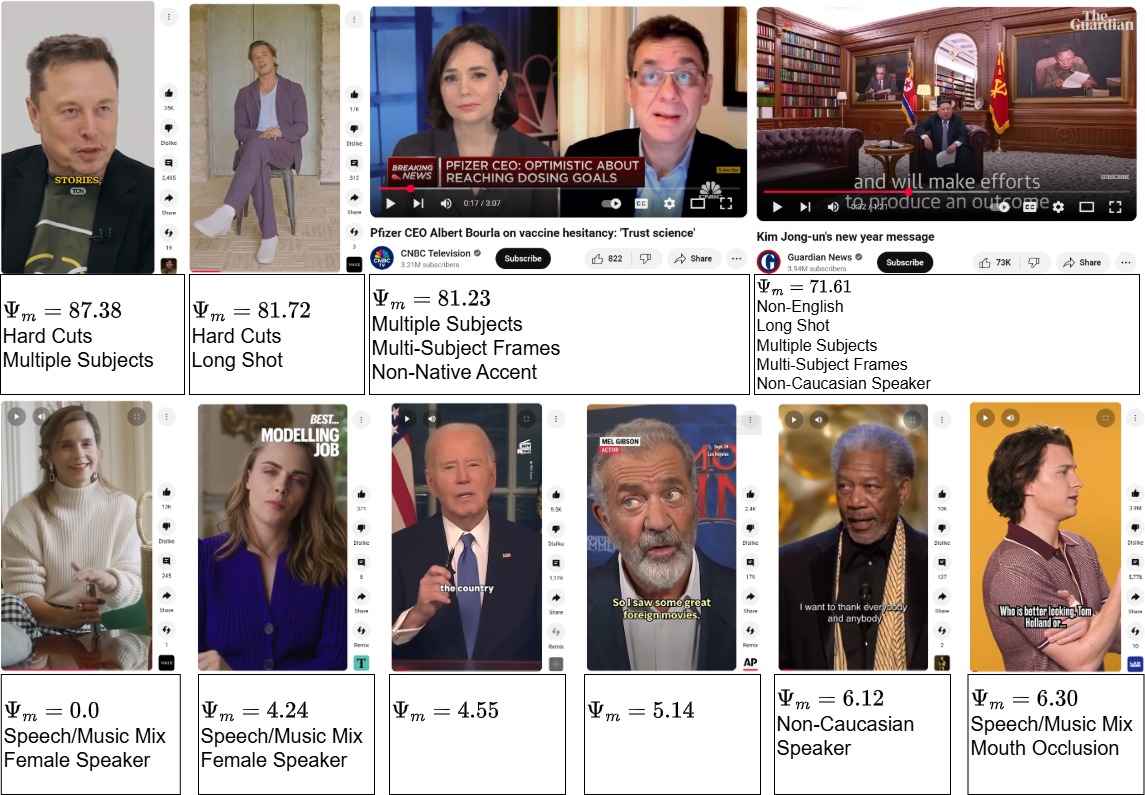}
    \caption{Examples of erroneously (upper panel; 1: \url{https://www.youtube.com/shorts/c15HhmtfI5w}, 2: \url{https://www.youtube.com/shorts/cqxkYUfOf9c}, 3: \url{https://www.youtube.com/watch?v=P6O6FqYYImk&ab_channel=CNBCTelevision}, 4: \url{https://www.youtube.com/watch?v=YLWqqDVWsXo}) and correctly (bottom panel; 1: \url{https://www.youtube.com/shorts/ciSd7M3lAco}, 2: \url{https://www.youtube.com/shorts/kyf_eCCc5qU}, 3: \url{https://www.youtube.com/shorts/EVQo--qKQAI}, 4: \url{https://www.youtube.com/shorts/dvfN3WD1_A8},  5: \url{https://www.youtube.com/shorts/JW4FT8RVJ88}, 6: \url{https://www.youtube.com/shorts/gH96b2U293Y}) classified \textbf{real} samples.}
    \label{fig:real_examples}
\end{figure*}

\Cref{fig:real_examples} provides further qualitative examples pinpointing the most prominent contributing factors to the performance drop from a controlled to the real-world uncontrolled environment.

\end{document}